\pdfoutput=1
\documentclass[11pt,a4paper]{article}
\usepackage[hyperref]{acl2020}
\usepackage{times}
\usepackage{latexsym}

\usepackage{csquotes}
\usepackage{booktabs}
\usepackage{fancyvrb}
\usepackage{graphicx}

\usepackage{microtype}

\aclfinalcopy 

\title{Is 42 the Answer to Everything \\
in Subtitling-oriented Speech Translation?}

\author{Alina Karakanta \\
  Fondazione Bruno Kessler \\
  University of Trento \\
  Trento - Italy \\
  \texttt{akarakanta@fbk.eu} \\\And
  Matteo Negri \\
  Fondazione Bruno Kessler \\
  Trento - Italy \\
  \texttt{negri@fbk.eu} \\\And
  Marco Turchi \\
  Fondazione Bruno Kessler \\
  Trento - Italy \\
  \texttt{turchi@fbk.eu} \\}

\date{}

\begin{document}
\maketitle
\begin{abstract}
Subtitling is becoming increasingly important for disseminating information, given the enormous amounts of audiovisual content becoming available daily. Although Neural Machine Translation (NMT) can speed up the process of translating audiovisual content, large manual effort is still required for transcribing the source language, and for spotting and segmenting the text into proper subtitles. Creating proper subtitles in terms of timing and segmentation highly depends on information present in the audio (utterance duration, natural pauses). In this work, we explore two methods for applying Speech Translation (ST) to subtitling: a) a direct end-to-end and b) a classical cascade approach. 
 We discuss the benefit of having access to the source language speech for improving the conformity of the generated subtitles to the spatial and temporal subtitling constraints and show that length\footnote{Speaking of subtitles and their optimum length of 42 characters per line, we could not help but alluding to the book and series \textit{The Hitchhiker's Guide to the Galaxy} by Douglas Adams, where the number 42 is the ``Answer to the Ultimate Question of Life, the Universe, and Everything'', calculated by a massive supercomputer named Deep Thought for over 7.5 million years.} is not the answer to everything in the case of subtitling-oriented ST.
\end{abstract}

\section{Introduction}
Vast amounts of audiovisual content are becoming available every minute. From films and TV series, informative and marketing video material, to home-made videos, audiovisual content is reaching viewers with various needs and expectations, speaking different languages, all across the globe. This unprecedented access to information through audiovisual content is made possible mainly thanks to subtitling. Subtitles, despite being the fastest and most wide-spread way of translating audiovisual content, still rely heavily on human effort. In a typical multilingual subtitling workflow, a subtitler first creates a subtitle template \cite{georgakopoulou-19} by transcribing the source language audio, timing and adapting the text to create proper subtitles in the source language. These source language subtitles (also called captions) are already compressed and segmented to respect the subtitling constraints of length, reading speed and proper segmentation \cite{Avt2007sub,karakanta2019subtitling}.
In this way, the work of an NMT system is already simplified, since it only needs to translate matching the length of the source text \cite{matusov-etal-2019-customizing,lakew-et-al-19-len}. However, the essence of a good subtitle goes beyond matching a predetermined length 
(as, for instance, 42 characters per line in the case of TED talks). Apart from this spatial dimension, subtitling relies heavily on the temporal dimension, which is incorporated in the subtitle templates in the form of timestamps. However, templates are expensive and slow to create and as so, not a viable solution for short turn-around times and individual content creators. Therefore, skipping the template creation process would greatly extend the application of NMT in the subtitling process, leading to massive reductions in costs and time and making multilingual subtitling more accessible to all types of content creators.

In this work, we propose Speech Translation as an alternative to the template creation process. 
We experiment with cascade systems, 
i.e. pipelined ASR+MT architectures, and 
direct, end-to-end ST systems. While the MT system in the pipelined approach receives a raw textual transcription as input, the direct speech translation receives temporal and prosodic information from the source language input signal. Given that several decisions about the form of subtitles depend on the audio (e.g. subtitle segmentation at natural pauses, length based on utterance duration), in this work we ask the question whether end-to-end ST systems can take advantage of this information to better model the subtitling constraints and subtitle segmentation. In other words, we investigate whether, in contrast to text translation (where one mainly focuses on returning subtitles of a maximum length of 42 characters), in speech translation additional information can help to develop a more advanced approach that goes beyond merely matching the source text length.

Our contributions can be summarised as follows:
\begin{itemize}
    \item We present the first end-to-end solution for subtitling, completely eliminating the need of a source language transcription and any extra segmenting component;
    \item We conduct a thorough analysis of cascade vs. end-to-end systems both in terms of translation quality and conformity to the subtitling contraints, showing that end-to-end systems have large potential for subtitling.
\end{itemize}

\section{Background}

\subsection{Subtitling}\label{sec:subtitling}
Subtitling involves translating the audio (speech) in a video into text in another language (in the case of interlingual subtitling). Subtitling is therefore a multi-modal phenomenon, incorporating visual images,  gestures,\footnote{While we recognise the importance of the visual dimension in the process of subtitling, incorporating visual cues in the NMT system is beyond the scope of this work.} sound and language \cite{Taylor-2016}, but also an intersemiotic process, in the sense that it involves a change of channel, medium and code from speech to writing, from spoken verbal language to written verbal language \cite{assis-2001}.

The temporal dimension of subtitles and the relation between audio and text has been stressed also in professional subtitling guidelines. In their subtitling guidelines, \citet{Carroll-Ivarsson-98} mention that:
\textit{``The in and out times of subtitles must follow the speech rhythm of the dialogue, taking cuts and sound bridges into consideration''} and that: \textit{``There must be a close correlation between film dialogue and subtitle content; source language and target language should be synchronized as far as possible''}. Therefore, a subtitler's decisions are guided not only by attempting to transfer the source language content, but also by achieving a high correlation between the source language speech and target language text.

In addition, subtitles have specific properties in the sense that they have to conform to spatial and temporal constraints in order to ensure comprehension and a pleasant user experience. For example, due to limited space on screen, a subtitle cannot be longer than a fixed number of characters per line, ranging between 35-43 \cite{Avt2007sub}
. When it comes to the temporal constraints, a comfortable reading speed (about 21 chars/second) is key to a positive user experience. It should be ensured that viewers have enough time to read and assimilate the content, while at the same time their attention is not monopolised by the subtitles
. Lastly, the segmentation of subtitles should be performed in a way that facilitates comprehension, by keeping linguistic units
(e.g. phrases) in the same line.
For the reasons mentioned above, subtitlers should ideally always have access to the source video when translating. However, working directly from the video can have several drawbacks. The subtitler needs to make sense of what is said on the screen, deal with regional accents, noise, unintelligible speech etc. 

One way to automatise this labour-intensive process, especially in settings 
where several target languages are involved, is creating a subtitle template of the source language \cite{georgakopoulou-19}. A subtitle template is an enriched transcript of the source language speech where the text is already compressed, timed and segmented into proper subtitles. This template can serve as a basis for translation into other languages, whether the translation is performed by a human or an MT system. 

In the case of templates, optimal duration, length and proper segmentation are ensured, since the change of code between oral and written in the source language has already been curated by an expert. Due to the high costs and time required, creating a subtitle template is not a feasible solution both for small content creators and in the case of high volumes and fast turn-around times. 

In the absence of a subtitle template, 
an automatic transcription of the source language audio could seem an efficient alternative. However, Automatic Speech Recognition (ASR) systems produce word-for-word transcriptions of the source speech, not adapted to the subtitling constraints, and where all information coming from the speech is discarded. This purely textual transcription is then translated by the MT system. Therefore, it is highly probable that a higher post-editing effort is required not only in terms of translation errors, but chiefly for repairing the form of the subtitles.

Direct, end-to-end speech translation receives audio as input. Therefore, the model receives two types of information from the spectrogram: 1) information about the temporal dimension of the speech, e.g. duration, and 2) information related to the frequency, such as pitch, power and other prosodic elements. While intonation and stress are mostly related to semantic properties, speech tempo directly affects the compression rate of subtitles, and pauses often correspond to prosodic chunks which can determine the subtitle segmentation. Given that, it is worth asking the question whether access to this information can lead to better modelling of the subtitling constraints and subtitle segmentation.

\subsection{Speech translation}
Traditionally, the task of Speech-to-Text Translation has been addressed with cascade systems consisting of two components: an ASR system, which transcribes the speech into text, and an MT system, which translates the transcribed text into the target language \cite{eck-2005}. This approach has the benefit that it can take advantage of state-of-the-art technology for both components and leverage the large amount of data available for both tasks. On the other hand, it suffers from error propagation from the ASR to the MT, since transcription errors are impossible to recover because the MT component typically does not have access to the audio. Several works have attempted to make MT robust to ASR errors \cite{digangi-2019-robust,sperber-2017}  
by working on noisy transcripts. 

One further drawback of the cascaded approach, particularly relevant for the task of subtitling, is that any transcript, no matter how accurate, is subject to information loss in the semiotic shift from the richer audio representation to the poorer text representation. This limitation has been addressed in the past in speech-to-speech translation cascades chiefly for improving the naturalness of the synthesised speech and for resolving ambiguities. This has been performed through acoustic feature vectors related to different prosodic elements, such as duration and power \cite{kano13interspeech}, emphasis \cite{do15iwslt,do16isattention} and intonation \cite{aguero06prosody,Anumanchipalli12Intent}.

By avoiding intermediate textual representations, end-to-end speech translation \cite{berard-2016} can cope with the above limitations. However, its performance and suitability for reliable applications has been impeded by the limited amount of training data available. In spite of this data scarcity problem, it has been recently shown that the gap between the two approaches is closing \cite{niehues-et-al18,niehues-et-al19}, especially with specially-tailored architectures \cite{digangi-2019-stransformer,dong-et-al-2018} and via effective data augmentation strategies \cite{Jia-2019-weakly}.
Despite an increasing amount of works attempting to improve the performance of ST for general translation
, there has been almost no work on comparing the two technologies on specific problems and applications, which is among the focus points of this work.

\subsection{Machine Translation for subtitling}
Despite the relevance of developing automatic solutions for subtitling both for the industry and academia, there have been very limited attempts to customise MT for subtitling. Previous works based on Statistical MT (SMT) used mostly proprietary data and led to completely opposite outcomes. \citet{Volk-et-al-2010} developed SMT systems for the Scandinavian TV industry and reported very good results in this practical application. \citet{aziz-etal_EAMT:2012} reported significant reductions in post-editing effort compared to translating from scratch for DVD subtitles for English-Portuguese. On the other hand, the SUMAT project \cite{bywood-2013-sumat, bywood-2017-sumat}, involving seven European language pairs, concluded that subtitling poses particular challenges for MT and therefore a lot of work is still required before MT can lead to real improvements in audiovisual translation workflows \cite{burchardt-16}.

Recently, after the advent of the neural machine translation paradigm, \citet{matusov-etal-2019-customizing} presented an NMT system customised to subtitling. The main contribution of the paper is a segmenter module trained on human segmentation decisions, which splits the resulting translation into subtitles.
The authors reported reductions in post-editing effort, especially regarding subtitle segmentation. On a different strand of research, \citet{lakew-et-al-19-len} proposed two methods for controlling the output length in NMT. The first one is based on adding a token, as in Multilingual NMT \cite{johnson-etal-2017-googles,Ha-2016-multi}, which in this setting represents the length ratio between source and target, and the second inserts length information in the positional encoding of the Transformer. 

The application of MT (either SMT or NMT) in the works described above is possible only because of the presence of a ``perfect'' source language transcript, either for the translation itself or for computing the length ratio. To our knowledge today, no work so far has experimented with direct end-to-end ST in the domain of subtitling.

\section{Experimental Setup}
\label{sec:setup}

\subsection{Data}\label{sec:data}
For the experiments we use the MuST-Cinema corpus \cite{mustcinema20},\footnote{Must-Cinema has been derived from the MuST-C corpus \cite{di_gangi_must-c:_2019}, which currently represents the largest multilingual corpus for ST.} which contains (\textit{audio}, \textit{transcription}, \textit{translation}) triplets 
where the breaks between subtitles have been annotated with special symbols. The symbol \texttt{<eol>} corresponds to a line break inside a subtitle block, while the symbol \texttt{<eob>} to a subtitle block break (the next subtitle comes on a different screen), as seen in the following example from the MuST-Cinema test set: 

\textit{  This kind of harassment keeps women \texttt{<eol>} from accessing the internet --
\texttt{<eob>} \\ essentially, knowledge. \texttt{<eob>}}

We experiment with 2 language pairs, English$\rightarrow$French and English$\rightarrow$German, as languages with different syntax and word order. The training data consist of 229K and 275K sentences (408 and 492 hours) for German and French respectively, while the development sets contain 1088/1079 sentences and the test sets 542/544 sentences.

\subsection{MT and ST systems}
The \textbf{Cascade} system consists of an ASR and an MT component. 
The ASR component is based on the KALDI toolkit \cite{povey2011kaldi}, featuring a time-delay neural network and lattice-free maximum mutual information discriminative sequence-training ~\cite{povey2016}. The audio data for acoustic modelling include the clean portion of LibriSpeech \citep{librispeech} ($\sim$460h) and a variable subset of the MuST-Cinema training set ($\sim$450h), from which 40 MFCCs per time frame were extracted. A MaxEnt language model~\cite{Alumae2010} is estimated from the corresponding transcripts ($\sim$7M words). 
The MT component is based on the Transformer architecture (big)~\cite{vaswani2017attention} with similar settings to the original paper. The system is first trained on the OPUS data, with 120M sentences for EN$\rightarrow$FR and 50M for EN$\rightarrow$DE and then fine-tuned on MuST-Cinema. 
Considering that the ASR output is lower-cased and without punctuation, we lowercase and remove the punctuation from the source side of the parallel data used in pre-training the MT system. To mitigate the error propagation between the ASR and the MT, for fine-tuning, we use a version of MuST-Cinema where the source audio has been transcribed by the tuned ASR.

For the \textbf{End-to-End} system, we experiment with two data conditions, one where we only use the MuST-Cinema training data (\textbf{E2E-small}) and a second one where we pre-train on a larger amount of data and fine-tune on MuST-Cinema (\textbf{E2E}). This will allow us to detect whether there is any trade-off between translation quality and conformity to constraints when increasing the amount of training data that are not representative of the target application (subtitling). The architecture used is S-Transformer, \cite{digangi-2019-stransformer}, an ST-oriented adaptation of Transformer, which has been shown to achieve high performance on different speech translation benchmarks. We remove the 2D self-attention layers and increase the size of the encoder to 11 layers, while for the decoder we use 4 layers. This choice was motivated by preliminary experiments, where we noted that replacing the 2D self-attention layers with normal self-attention layers and adding more layers in the encoder increased the final score, while removing a few decoder layers did not negatively affect the performance. As distance penalty, we choose the logarithmic distance penalty. 
We use the encoder of the ASR model to initialise the weights of the ST encoder and achieve faster convergence \cite{bansal-etal-2019-pre}. 

Since the E2E-small system, trained only on MuST-Cinema, is disadvantaged in terms of the amount of training data compared to the cascade, we utilise synthetic data to boost the performance of the ST system (E2E). 
To this aim, we automatically translate into German and into French the English transcriptions of the data available for the IWSLT2020 offline speech translation task\footnote{\url{http://iwslt.org/doku.php?id=offline_speech_translation}} (whenever the translation is not available in the respective target language). To this aim, we use an MT Transformer model achieving 43.2 BLEU points on the WMT'14 test set \cite{ott-etal-2018-scaling} for EN$\rightarrow$FR. Our EN$\rightarrow$DE Transformer model using similar settings achieves 25.3 BLEU points on the WMT'14 test set. 
The resulting training data (both real and synthetic) amount to 1.5M sentences on the target side. We use a different tag to separate the real from the synthetic data. We further use SpecAugment \cite{Park_2019}, a technique for online data augmentation, with augment rate of 0.5. Finally, we fine-tune on MuST-Cinema. 

For comparison, we also report MT results when starting from a ``subtitle template'' (\textbf{Template}). In this setting, we use the textual source side of MuST-Cinema, which contains the human transcriptions of the source language audio and its segmentation into subtitles. In this way, the input to the MT system is already split in subtitles using the special symbols, respecting the subtitling constraints and with proper segmentation.  
This will allow us to have an upper-bound of the performance that NMT can achieve when provided with input already in the form of subtitles. We pre-train large models with the OPUS data used in the cascade but without lowercasing or removing punctuation and then fine-tune them on the full training set of MuST-Cinema. It should be noted that only the MuST-Cinema data contain break symbols. We use the same Transformer architecture as in the cascade system.  
For all the experiments we use the fairseq toolkit \cite{gehring-etal-2017-convolutional}. Models are trained until convergence. Byte-Pair Encoding (BPE) \cite{sennrich-etal-2016-neural} is set to 8K operations for the E2E-small and E2E systems while to 50K joint operations for the Cascade and the Template systems.

\subsection{Evaluation}
To evaluate translation quality we use BLEU \cite{papineni-etal-2002-bleu} against the MuST-Cinema test set, both with the break symbols and after removing them (BLEU-nob). For BLEU, an incorrect break symbol would account for an extra $n$-gram error in the score computation, while BLEU-nob allows us to evaluate only the translation quality without taking into account the subtitle segmentation. 

For evaluating the conformity to the constraint of length, we calculate the percentage of subtitles with a maximum length of 42 characters per line (CPL), while for reading speed the percentage of sentences with maximum 21 characters per second (CPS). Since the MuST-Cinema data come from TED talks, these values were chosen according to the TED subtitling guidelines\footnote{\url{https://translations.ted.com/TED_Translator_Resources:_Main_guide}}.

Finally, for judging the goodness of the segmentation, i.e. the position of the breaks in the translation, we mask all words except for the break symbols and compute Translation Edit Rate  \cite{Snover06astudy} only for the breaks against the reference translation (TER-br). This will allow us to determine the effort required by a human subtitler to manually correct the segmentation.

\section{Results}\label{sec:results}
\subsection{Translation quality}
The results are shown Table~\ref{tab:results}. 
\begin{table*}[t]
\small
    \centering
    \begin{tabular}{ll|lcccc} \toprule
         & & BLEU$\uparrow$ & BLEU-nob$\uparrow$ & CPL$\uparrow$ & CPS$\uparrow$ & TER-br$\downarrow$ \\ \midrule
        FR & Template & 30.62 & 28.86 & 91$\%$ & 68$\%$ & 18\\
        & Cascade & 22.41$\dagger$	& 22.06 & 93$\%$ & 72$\%$ & 22 \\
        & E2E-small & 18.76 & 18.03 & 95$\%$ & 70$\%$ & 23 \\
        & E2E & 22.22$\dagger$ & 21.9 & 95$\%$ & 70$\%$ & 20 \\ \midrule
        DE & Template & 22.08 & 21.10 & 90$\%$ & 56$\%$ & 17\\
        & Cascade & 17.81$\dagger$ & 17.82 & 90$\%$ & 56$\%$ & 21 \\
        & E2E-small & 11.92 & 11.38 & 93$\%$ & 55$\%$ & 24 \\
        & E2E & 17.28$\dagger$ & 16.90 & 92$\%$ & 56$\%$ & 20 \\ \bottomrule
    \end{tabular}
    \caption{Results for translation quality (BLEU, BLEU-nob), for conformity to the subtitling constraints (CPL, CPS) and for subtitle segmentation (TER-br) for the four systems. 
    Results marked with $\dagger$ are not statistically significant.
    }
    \label{tab:results}
\end{table*}
As far as translation quality is concerned, the best performance is reached, as expected, in the Template setting, where the MT system is provided with 
``perfect'' source language transcriptions. On the MuST-Cinema test set, this leads to BLEU scores of 30.62 and 22.08 respectively for French and German.
The Cascade setting follows with 
a BLEU score reduction
of 8 points for French and 4 points for 
German, which can be attributed to  error propagation from the ASR component to the MT.

The E2E-small model (trained solely on MuST-Cinema) achieves 18.76 and 11.92 BLEU points for French and German respectively, which is a relatively low performance compared to the rest of the systems. This ``low-resource'' setting is a didactic experiment aimed at exploring how far the data-hungry neural approach can go with the limited amount of data available in the domain of ST for subtitling (280K and 234K sentences). It should be also noted that MuST-Cinema is the only Speech Translation corpus of the subtitle genre, both respecting the subtitling constraints and containing break symbols. Consequently, the interference of other data may hurt the conformity to the subtitling constraints, despite improving the translation performance. On the other hand, pre-training has evolved in a standard procedure for coping with the data-demanding nature of NMT. Therefore, pre-training also in the case of end-to-end speech translation offers a comparable setting with the template and the cascade experiments. Indeed, after fine-tuning the pre-trained model on MuST-Cinema, E2E reaches 22.22 and 17.28 BLEU points for French and German. The difference in translation quality is
not statistically significant between the Cascade and the E2E, with the Cascade scoring higher with 0.2/0.6 BLEU points for French/German respectively. 
This shows that when increasing the size of the training data for the E2E 
the gap between the cascade and the end-to-end approach is closed and that end-to-end approaches may have finally found a steady ground for flourishing in different applications.

\subsection{Conformity to the subtitling constraints and subtitle segmentation}
When it comes to the conformity to the subtitling constraints, the results show a different picture 
(see CPS and CPL of Table \ref{tab:results}). E2E exceeds all models at achieving proper length of subtitles, with 95\% and 93\% of the subtitles having length of maximum 42 characters. E2E achieves higher conformity with length even compared to the Template, for which the segmentation is already provided to the system in the form of break symbols, while the cascade is behind by 2\%. The same tendency is observed in the TER-br results computed to measure the proper placement of the break symbols. While the Template benefits from the source language segmentation and therefore requires less edits to properly segment the subtitles, the Cascade is disadvantaged in guessing the correct position of the break symbols, as shown by a 22 and 21 TER score. For E2E-small TER-br is higher, possibly due to the low translation quality. However, E2E outperforms the Cascade by 1 TER point in this respect, showing that less effort would be required to segment the sentences into subtitles.
This suggests that the E2E system receives information compared to the Cascade, which allows for better guessing the positions of the break symbols in the translation. This is another indication that subtitle segmentation decisions are not solely determined by reaching a maximum length of 42 characters, but a combination of multiple (possibly intersemiotic) factors can offer a better answer to automatic subtitling.

\section{Analysis}
The higher scores for CPL and TER-br in Section~\ref{sec:results} suggest that the E2E system is better at modelling the subtitling constraints of length and proper segmentation. In this section we shed more light into this aspect by analysing factors which might be determining the system's behaviour in relation to the insertion of the break symbols \texttt{<eol>} and \texttt{<eob>}. 

One question quickly arising is how the system can determine whether to insert a subtitle break symbol \texttt{<eob>} (which means that the next subtitle will follow on a new screen) or a line break symbol \texttt{<eol>} (which means that the next line of the subtitle will appear on the same screen). Since the maximum number of lines allowed per subtitle block is 2, a simple answer would be to alternate between   \texttt{<eob>} and  \texttt{<eol>} such that all subtitles would consist of two lines (two-liners). However, anyone having watched a film with subtitles is aware that subtitles can be two-liners or one-liners. Coming back to the example in Section~\ref{sec:data}, depending on the choice of break symbols (except for the last symbol which should always be an \texttt{<eob>}), there are two possible renderings of the subtitle:

\begin{Verbatim}[fontsize=\small]
10
00:00:31,066 --> 00:00:34,390
This kind of harassment keeps women
from accessing the internet --
11
00:00:34,414 --> 00:00:36,191
essentially, knowledge.
\end{Verbatim}

and 

\begin{Verbatim}[fontsize=\small]
10
00:00:31,066 --> 00:00:34,390
This kind of harassment keeps women
11
00:00:34,414 --> 00:00:36,191
from accessing the internet --
essentially, knowledge.
\end{Verbatim}

Only the first rendering is acceptable because it satisfies the reading speed constraint but also corresponds to the speech rhythm, since the speaker makes a pause after uttering the word ``internet''. In this case, how can the MT system determine which type of break symbol to insert?

We have mentioned in Section~\ref{sec:subtitling} that the \textit{in} and \textit{out} times of a subtitle should follow the rhythm of speech. Therefore, we expect that the end of a subtitle block, which in our setting is signalled by the break symbol \texttt{<eob>}, should correspond to the end of a speech act, a pause or a terminal juncture. On the other hand, line breaks inside the same subtitle block, which in our work correspond to the break symbol \texttt{<eol>}, have a different role. While line breaks can still overlap with pauses or signal the change of speaker, their function is to split a long subtitle into two smaller parts in order to fit the screen. The decision of where to insert a line break inside a subtitle block is determined by two factors: achieving a more or less equal length of the upper and the lower subtitle line and inserting the break in a position such that syntactic units are kept together. Consequently, the insertion of \texttt{<eol>} is determined more by the length and the syntactic properties of the subtitle and less by the natural rhythm of the speech.

If the hypothesis above holds, the choice of whether to insert an \texttt{<eob>} or an \texttt{<eol>} symbol is defined by prosodic properties and not solely by reaching the maximum length of 42 characters. As a consequence, it is not a simple alternating procedure. 

To test this hypothesis, we compute the duration of the pause coming after each word in the source side of the MuST-Cinema test set. To achieve this, we perform forced alignment of the transcript against the audio 
and subtract the end time of each word from the start time of the next word:
\begin{equation}
    pause_{w1w2} = start\_time_{w2} - end\_time_{w1}
\end{equation}

Then we separate the pauses in 3 groups: \textit{i)} pauses corresponding to positions where \texttt{<eob>} is present, \textit{ii)} pauses corresponding to positions where \texttt{<eol>} is present and \textit{iii)} pauses after which there is no break symbol (\textit{None}). 
In Table~\ref{tab:pauses} we report average and standard deviation of the pause duration for each category.

\begin{table}[h]
\small
    \centering
    \begin{tabular}{c|cc}\toprule
    Pause type & Avg & Stdev \\\midrule
    None & 0.039 & 0.022 \\
    \texttt{<eob>} & 0.551 & 0.181 \\
    \texttt{<eol>} & 0.074 & 0.027 \\ \bottomrule
    \end{tabular}
     \caption{Average pause duration and standard deviation (in seconds) for the category without breaks (None), and for the categories with the two types of break symbols \texttt{<eob>} and \texttt{<eol>}.}

    \label{tab:pauses}
\end{table}

\begin{table*}[t]
\small
    \centering
    \begin{tabular}{l|c}\toprule
EN & ``Who do you report to?'' \texttt{<eob>} ``It depends''. 
    \\    
CS & Wen melden Sie an? \texttt{<eol>} Es hängt davon ab.  \\
E2E & Wen berichten Sie? \texttt{<eob>} Es kommt darauf an.  \\
REF & ``An wen schickst du deine Berichte?'' \texttt{<eob>} ``Das kommt darauf an''. \\
 \midrule
\multicolumn{2}{c}{One executive at another company \texttt{<eob>} likes to explain how he used to be \texttt{<eol>} a master of milestone-tracking.} \\
\multicolumn{2}{c}{Un cadre d'une autre entreprise aime expliquer \texttt{<eob>} comment il était jadis un maître \texttt{<eol>} du trek capital.}  \\
\multicolumn{2}{c}{Un dirigeant d'une autre entreprise \texttt{<eob>} aime expliquer comment il était \texttt{<eol>} un maître de traçage en pierre.} \\
\multicolumn{2}{c}{Un cadre d'une autre entreprise \texttt{<eob>} aime raconter comme il était passé maître \texttt{<eol>} dans la surveillance des étapes.}  \\
\midrule
\multicolumn{2}{c}{But you know how they say \texttt{<eol>} that information is a source of power?} \\
\multicolumn{2}{c}{Vous savez comment dire que l'information \texttt{<eol>} est une source de pouvoir.} \\
\multicolumn{2}{c}{Saviez-vous comment dire \texttt{<eol>} que l'information est une source de pouvoir ?}  \\
\multicolumn{2}{c}{Mais vous savez qu'on dit que \texttt{<eol>} l'information est source de pouvoir ? } \\
 \bottomrule
    \end{tabular}
    \caption{Examples of translations by the cascade (CS) and the end-to-end model (E2E) compared to the source sentence (EN) and the reference (REF). The sentence-final \texttt{<eob>} has been removed.}
    \label{tab:examples}
\end{table*}

Pauses corresponding to the positions where \texttt{<eob>} symbols are present are more than x10 longer than the pauses in positions without any break symbols (None). Even if we take the most extreme cases (based on standard deviation), any pause above 0,37 seconds requires the insertion of \texttt{<eob>}. Pauses corresponding to \texttt{<eol>} symbols are on average x2 longer, but there is an overlap between the possible durations of the \textit{None} and the \texttt{<eol>} category. This confirms our hypothesis about the different roles of the two subtitle breaks. Therefore, prosodic information is an important factor which can help the ST system determine the subtitling segmentation according to the speech rhythm.
This finding provides strong evidence towards a clear limitation of the cascade setting, where the raw textual transcription from the ASR does not provide any prosodic information to the MT system. 
The MT system in the cascade setting is disadvantaged by the inability to: \textit{i)} recover from possible ASR errors, and \textit{ii)} make decisions determined by factors other than text.

With this knowledge, we analyse the breaks in the results of the two systems. In order to control for differences in translation, we select all sentences with at least 2 breaks and with the same number of break symbols (regardless of whether \texttt{<eob>} or \texttt{<eol>}) between the reference, the output of the Cascade and of the E2E. The resulting sentences are 137 for French and 158 for German. We calculate the accuracy of the type of break symbols for the two systems. For French the accuracy is 89\% for the Cascade and 93\% for the E2E. For German the accuracy is 85\% for the Cascade and 88\% for the E2E. This difference in accuracy suggests that the E2E is aided by the acoustic information and specifically by the pause duration in determining the correct break symbol.

Table~\ref{tab:examples} presents some examples, evaluated also against the video. In the first example, the decision of which type of break to insert between the two sentences can only be determined by the duration of the pause that comes between them. Indeed, the speaker in the 
video asks the question and then leaves some time to the audience before giving the answer. The pause between the two sentences is about 2 seconds. In this case, the first sentence should be in one subtitle block, then disappear, and the second sentence should come in the next subtitle block in order not to reveal the answer before it was spoken by the speaker. This information is only available to the E2E system. 

In the second example, although both systems have chosen the right type of break, there are differences in the actual positions of the breaks. The E2E inserts the first break symbol in the same position as in the source and reference (\textit{entreprise}), while the Cascade inserts it at a later position (\textit{expliquer}), resulting in a subtitle of 46 characters, which is above the 42-character length limit. The Cascade correctly inserts the second break (\textit{maître}), as in the reference. Here, the E2E copies the break position from the source sentence, which is in a different position compared to the reference (after the word \textit{be} instead of the word \textit{master} as in the reference). The E2E is faithful to the segmentation of the source language when it corresponds to the pauses of the speaker. The positions chosen by the Cascade to insert the break symbols are before a conjunction (\textit{comment}) and a preposition (\textit{du}). Contrary to the E2E, the Cascade's decisions are based more on syntactic patterns, learned from the existing human segmentation decisions in the training data. 

The third example shows that prosody is important also for other factors related to the translation. The Cascade, not receiving any punctuation, was not able to reproduce the question in the translation, while the intonation might have helped the E2E to render the sentence as a question despite using the wrong tense (\textit{saviez} instead of \textit{savez}). 

All in all, these examples confirm our analysis and once again indicate the importance of considering the intersemiotic nature of subtitling when 
developing MT systems for this task.

\section{Conclusion}
We have presented the first Speech Translation systems specifically tailored for subtitling. The first system is an ASR-MT cascade, while the second a direct, end-to-end ST system. These systems allow, for the first time, to create satisfactory subtitles both in terms of translation quality and conformity to the subtitling constraints in the absence of a human transcription of the source language speech (template). We have shown that while the two systems have similar translation quality performance, the E2E seems to be modelling the subtitle constraints better. We show that this could be attributed to acoustic features, such as natural pauses, becoming available to the E2E system through the audio input. This leads to a segmentation closer to the speech rhythm, which is key to a pleasant user experience. Our work takes into account the intersemiotic nature of subtitling by avoiding conditioning the translation on the textual source language length, as in previous approaches to NMT for subtitling, arriving to the conclusion that 42 is not the answer to everything in the case of NMT for subtitling. Rather, key elements for good automatic subtitling are prosodic elements such as intonation, speech tempo and natural pauses. We hope that this work will pave the way for developing more comprehensive approaches to NMT for subtitling.

\section*{Acknowledgments}
This work is part of the ``End-to-end Spoken Language Translation in Rich Data Conditions'' project,\footnote{\url{https://ict.fbk.eu/units-hlt-mt-e2eslt/}} which is financially supported by an Amazon AWS ML Grant.

\bibliography{anthology,acl2020}
\bibliographystyle{acl_natbib}

\end{document}